\documentclass[smallcondensed,natbib]{svjour3}     
\RequirePackage{fix-cm}
\smartqed  
\usepackage{graphicx}
\usepackage{url}
\usepackage{mathtools}
\usepackage[flushleft]{threeparttable}
\usepackage{siunitx}
\usepackage{algorithm} 
\usepackage{algorithmic}
\usepackage{siunitx}
\usepackage{arydshln}
\usepackage{subfig}
\usepackage{multirow}
\usepackage{tabularx}

\usepackage{natbib}
\def\ie{\emph{i.e.}\,}
\def\eg{\emph{e.g.}\,}

 \journalname{myjournal}
\begin{document}

\title{EVE: Explainable Vector Based Embedding Technique Using Wikipedia
}


\author{M. Atif Qureshi       \and
        Derek Greene 
}


\institute{M. Atif Qureshi \at
              Insight Centre for Data Analytics, University College Dublin, Dublin, Ireland \\
              \email{muhammad.qureshi@ucd.ie}           
           \and
           Derek Greene \at
              Insight Centre for Data Analytics, University College Dublin, Dublin, Ireland \\
              \email{derek.greene@ucd.ie}
}

\date{}

\maketitle

\begin{abstract}
We present an unsupervised explainable word embedding technique, called \textit{EVE}, which is built upon the structure of Wikipedia. The proposed model defines the dimensions of a semantic vector representing a word using human-readable labels, thereby it readily interpretable. Specifically, each vector is constructed using the Wikipedia category graph structure together with the Wikipedia article link structure. To test the effectiveness of the proposed word embedding model, we consider its usefulness in three fundamental tasks: 1) intruder detection --- to evaluate its ability to identify a non-coherent vector from a list of coherent vectors, 2) ability to cluster --- to evaluate its tendency to group related vectors together while keeping unrelated vectors in separate clusters, and 3) sorting relevant items first --- to evaluate its ability to rank vectors (items) relevant to the query in the top order of the result. For each task, we also propose a strategy to generate a task-specific human-interpretable explanation from the model. These demonstrate the overall effectiveness of the explainable embeddings generated by \textit{EVE}. Finally, we compare \textit{EVE} with the \textit{Word2Vec}, \textit{FastText}, and \textit{GloVe} embedding techniques across the three tasks, and report improvements over the state-of-the-art. 
\end{abstract}

\section{Introduction}
\label{sec:intro}
Recently the European Union has approved a regulation which requires that citizens have a ``right to explanation'' in relation to any algorithmic decision-making \citep{eu-regulation-goodman2016}. According to this regulation, due to come into force in 2018, an algorithm that makes an automatic decision regarding a user, entitles that user to a clear explanation as to how the decision was made. With this in mind, we present an explainable decision-making approach to generating word embeddings, called the \textit{EVE} model. Word embeddings reference to a family of techniques that simply describes a concept (\ie word or phrase) as a vector of real numbers \citep{glove}.
These vectors have been shown useful in a variety of applications, such as topic modelling \citep{liu2015topical}, information retrieval \citep{diaz2016query}, and document classification \citep{kusner2015word}

Generally, word embedding vectors are defined by the context in which those words appear  \citep{baroni2014don}. Put simply, ``a word is characterized by the company it keeps" \citep{firth1957}. To generate these vectors, a number of unsupervised techniques have been proposed which includes applying  neural networks \citep{word2vec-cbow,word2vec-skipgram,fasttext}, constructing a co-occurrence matrix followed by dimensionality reduction \citep{levy2014neural,glove}, probabilistic models \citep{globerson2007euclidean,arora2016latent}, and explicit representation of words appearing in a context \citep{levy2014linguistic,levy2015improving}. 

Existing word embedding techniques do not benefit from the rich semantic information present in structured or semi-structured text. Instead they are trained over a large corpus, such as a Wikipedia dump or collection of news articles, where any structure is ignored. However, in this contribution we propose a model that uses the semantic benefits of structured text for defining embeddings. Moreover, to the best of our knowledge, previous word embedding techniques do not provide human-readable vector dimensions, thus are not readily open to human interpretation. In contrast, \textit{EVE} associates human-readable semantic labels with each dimension of a vector, thus making it an explainable word embedding technique. 

To evaluate \textit{EVE}, we consider its usefulness in the context of three fundamental tasks that form the basis for many data mining activities -- discrimination, clustering, and ranking. We argue for the need for objective evaluation-based strategies to ensure that subjective opinions are discouraged, which may be found tasks such as finding word analogies \citep{word2vec-cbow}. 
These tasks are applied to seven annotated datasets which differ in terms of topical content and complexity, where we demonstrate not only the ability of \textit{EVE} to successfully perform these tasks, but also its ability to generate meaningful explanations to support its outputs.


The reminder of the paper is organized as follows. In Section 2, we provide an overview of research relevant to this work. In Section 3, we provide background material covering the structure of Wikipedia, and then describe the methodology of the \textit{EVE} model in detail. In Section 4, we provide detailed experimental evaluation on the three tasks mentioned above, and also demonstrate the novelty of the \textit{EVE} model in generating explanations. Finally, in Section 5, we conclude the paper with further discussion and future directions. 
The relevant dataset and source code for this work can be publicly accessed at \url{http://mlg.ucd.ie/eve}.








\section{Related Work}
\label{sec:related}

 







Assessing the similarity between words is a fundamental problem in natural language processing (NLP). Research in this area has largely proceeded along two directions: 1) techniques built upon distributional hypothesis whereby contextual information serves as the main source for word representation; 2) techniques built upon knowledge bases whereby encyclopedic knowledge is utilized for determination of word associations. In this section, we provide an overview of these directions, along with a description of some works attempting to bridge the gap between techniques (1) and (2) above through knowledge-powered word embeddings. Finally, we conclude the section with an explanation of the novelty of \textit{EVE}.

\subsection{From Distributional Semantic Models to Word Embeddings}
\label{sec:rel-word-embed}
Traditional computational linguistics has shown the utility of contextual information for tasks involving word meanings, in line with the distributional hypothesis which states that ``linguistic items with similar distributions have similar meanings" \citep{harris1954distributional}. Concretely, distributional semantic models (DSMs) keep count-based vectors corresponding to co-occurring words, followed by a transformation of the vectors via weighting schemes or dimensionality reduction  \citep{baroni2010distributional,92hnc-contextvectors,schutze1992word}. A new family of methods, generally known as ``word embeddings", learns word representations in a vector space, where vector weights are set to maximize the probability of the contexts in which the word is observed in the corpus \citep{bengio2003neural,collobert2008unified}. 

A more recent type of word embedding technique \textit{word2vec} called into question the utility of deep models for learning useful representations, instead proposing continuous bag-of-words \citep{word2vec-cbow} and skip-gram \citep{word2vec-skipgram} models built upon a simple single-layer architecture. Another recent word embedding technique by \cite{glove} aims to combine best of both strategies, \ie usage of global corpus statistics available to traditional distributional semantics models and meaningful linear substructures. Finally, \cite{fasttext} proposed an improvement over \textit{word2vec} by incorporating character n-grams into the model, thereby accounting for sub-word information.







\subsection{Knowledge Base Approaches for Semantic Similarity and Relatedness}
\label{sec:rel-kb}
Another category of work which measures semantic similarity and relatedness between textual units relies on pre-existing knowledge resources (\eg thesauri, taxonomies or encyclopedias). Within the proposed works in the literature, the key differences lie in the knowledge base employed, the technique used for measurement of semantic distances, and the application domain \citep{Hoffart:2012:KKO:2396761.2396832}. Both \cite{budanitsky2006evaluating} and \cite{jarmasz2012roget} used generalization (`is a') relations between words using WordNet-based techniques; \cite{metzler2007similarity} used web search logs for measuring similarity between short texts, and both \cite{strube2006wikirelate} and \cite{gabrilovich2007computing} used rich encyclopedic knowledge derived from Wikipedia. \cite{witten2008effective} made use of tf.idf-like measures on Wikipedia links and \cite{yeh2009wikiwalk} made use of random walk algorithm over the graph driven from Wikipedia's hyperlink structure, infoboxes, and categories. Recently, \cite{jiang2015feature} utilize various aspects of page organizations within a Wikipedia article to extract Wikipedia-based feature sets for calculating semantic similarity between concepts. Also \cite{qureshi2015thesis} presented a Wikipedia-based semantic relatedness framework which uses Wikipedia categories and their sub-categories to a certain depth count to define the relatedness between two Wikipedia articles whose categories overlap with the generated hierarchies. 

\subsection{Knowledge-Powered Word Embeddings}
\label{sec:rel-kwe}
In order to resolve semantic ambiguities associated with text data, researchers have recently attempted to increase the effectiveness of word embeddings by incorporate knowledge bases when learning vector representations for words \cite{xu2014rc}. Two categories of works exist in this direction: 1) encoding entities and relations in a knowledge graph within a vector space with the goal of knowledge base completion\cite{bordes2011learning,socher2013reasoning}; 2) enriching the learned vector representations with external knowledge (from within a knowledge base) in order to improve the quality of word embeddings \cite{bian2014knowledge}. The works in the first category aim to train neural tensor networks for learning a d-dimensional vector for each entity and relation in a given knowledge base. The works in the second category leverage morphological and semantic knowledge from within knowledge bases as an additional input during the process of learning word representations.

The \textit{EVE} model relates to the works described in Section \ref{sec:rel-word-embed} in the sense that these models all attempt to construct word embeddings in order to characterize relatedness between words. However, like the approaches described in  Section \ref{sec:rel-kb}, \textit{EVE} also benefits from semantic information present in structured text, albeit with the different aim of producing word embeddings. The \textit{EVE} model is different from knowledge-powered word embeddings in that we produce a more general framework by learning vector representations for concepts rather than limiting the model to entities and/or relations. Furthermore, we utilize the structural organization of entities and concepts within a knowledge base to enrich the word vectors. A relevant recent work-in-progress, called \textit{ConVec} \citep{sherkat2017vector}, attempts to learn Wikipedia concept embeddings by making use of anchor texts (\ie linked Wikipedia articles). In contrast, \textit{EVE} gives a more powerful representation through the combination of Wikipedia categories and articles. Finally, a key characteristic that distinguishes \textit{EVE} from all existing models is its expressive mode of explanations, as enabled by the use of Wikipedia categories and articles.

\section{The \textit{EVE} Model}
\subsection{Background on Wikipedia}
Before we present the  methodology of the proposed \textit{EVE} model, we firstly provide background information on Wikipedia, whose underlying graph structure forms the basic building blocks of the model.

\begin{figure}[!t]
\begin{center}
\includegraphics[scale=0.37]{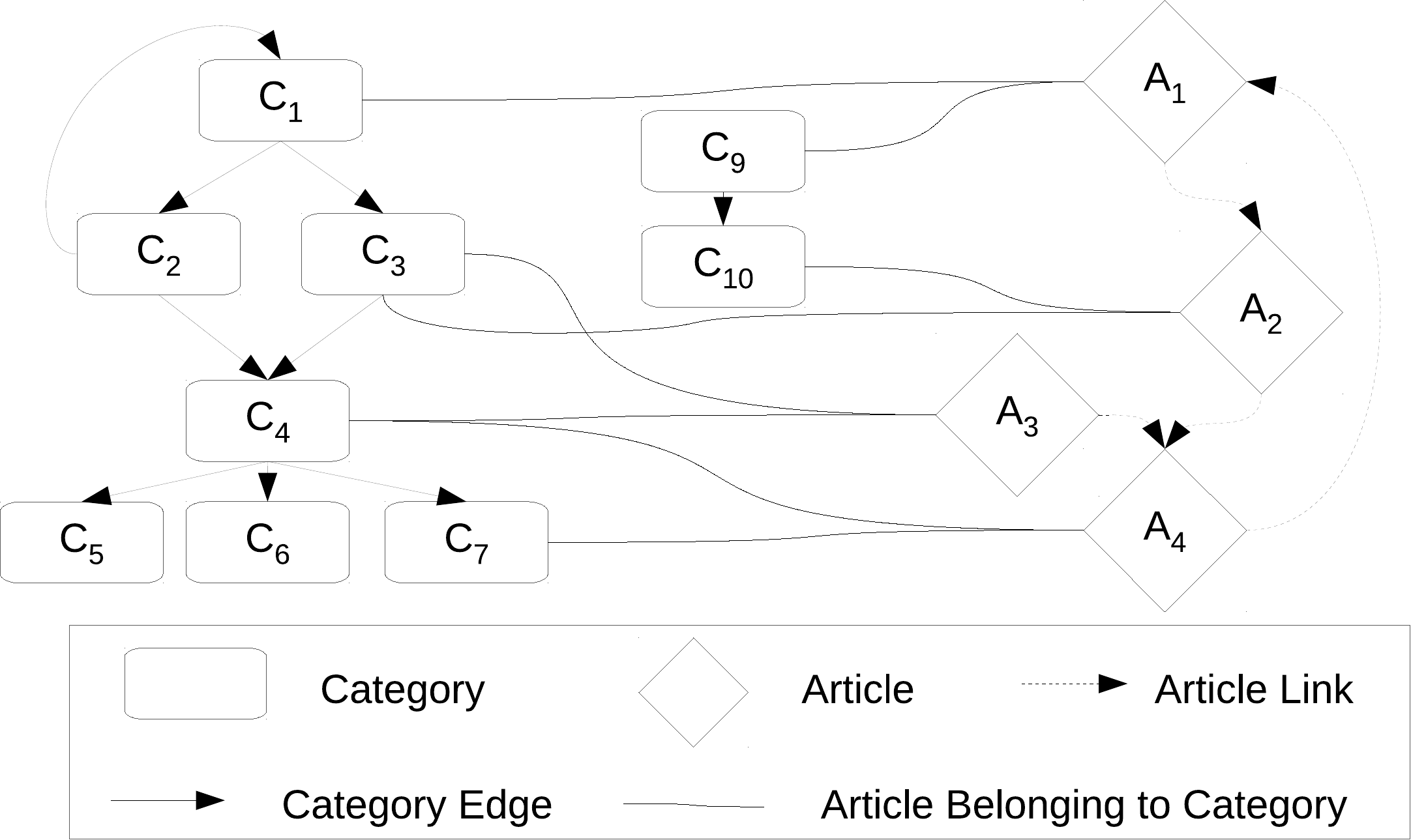}
\end{center}
\caption{An example Wikipedia graph structure for a set of four articles and ten associated categories.}
\label{fig:WCG}
\end{figure}

Wikipedia is a multilingual collaboratively-constructed encyclopedia  which is actively updated by a large community of volunteer editors. Figure \ref{fig:WCG} shows the typical Wikipedia graph structure for a set of articles and associated categories. Each article can receive an inlink from another Wikipedia article while it can also outlink to another Wikipedia article. In our example, article A\textsubscript{1} receives inlinks from A\textsubscript{4} and A\textsubscript{1} outlinks to A\textsubscript{2}. In addition, each article can belong to a number of categories, which are used to group together articles on a similar subject. In Fig. \ref{fig:WCG}, A\textsubscript{1} belongs to categories C\textsubscript{1} and C\textsubscript{9}. Furthermore, each Wikipedia category is arranged in a category taxonomy \ie, each category can have arbitrary number of super-categories and sub-categories. In our case, C\textsubscript{5}, C\textsubscript{6}, C\textsubscript{7} are sub-categories of C\textsubscript{4}, whereas C\textsubscript{2} and C\textsubscript{3} are super-categories of C\textsubscript{4}.

To motivate with a simple real example, the Wikipedia article ``Espresso'' receives inlinks from the article ``Drink'' and it outlinks to the article ``Espresso machine''. The article ``Espresso'' belongs to several categories, including ``Coffee drinks'' and ``Italian cuisine''. The category ``Italian cuisine'' itself has a number of super-categories (\eg ``Italian culture'', ``Cuisine by nationality'') and sub-categories (\eg ``Italian desserts'', ``Pizza''). These Wikipedia categories serve as a semantic tag for the articles to which they link \citep{zesch2007analysis}.

\subsection{Methodology}

We now present the methodology for generating word embedding vectors with the \textit{EVE} model. Firstly, a target word or concept is mapped to a single Wikipedia \emph{concept article}\footnote{This can be an exact match or a partial best match using an information retrieval algorithm}. 
The vector for this concept is then composed of two distinct types of dimensions. The first type quantifies the association of the concept with other Wikipedia articles, while the second type quantifies the association of the concept with Wikipedia categories. The intuition here is that related words or concepts will share both similar article link associations and similar category associations within the Wikipedia graph, while unrelated concepts will differ with respect to both criteria. The methods used to define these associations are explained next.

\subsubsection{Vector dimensions related to Wikipedia articles}
\label{sec:eve-article}
We firstly define the strategy for generating vector dimensions corresponding to individual Wikipedia articles. Given the target concept, which is mapped to a Wikipedia article denoted $A_{concept}$, we enumerate all incoming links and outgoing links between this article and all other articles. We then create a dimension corresponding to each of those linked articles, where the strength of association for a dimension is defined as the sum of the number of incoming and outgoing links involving an article and $A_{concept}$. After creating dimensions for all linked articles, we also add a \emph{self-link dimension}\footnote{This dimension the most relevant dimension defining the concept which is the article itself.}, where the association of $A_{concept}$ with itself is defined to be the twice of the maximum count received from the linking articles. 

\begin{figure}[!t]
\begin{center}
\includegraphics[scale=0.54]{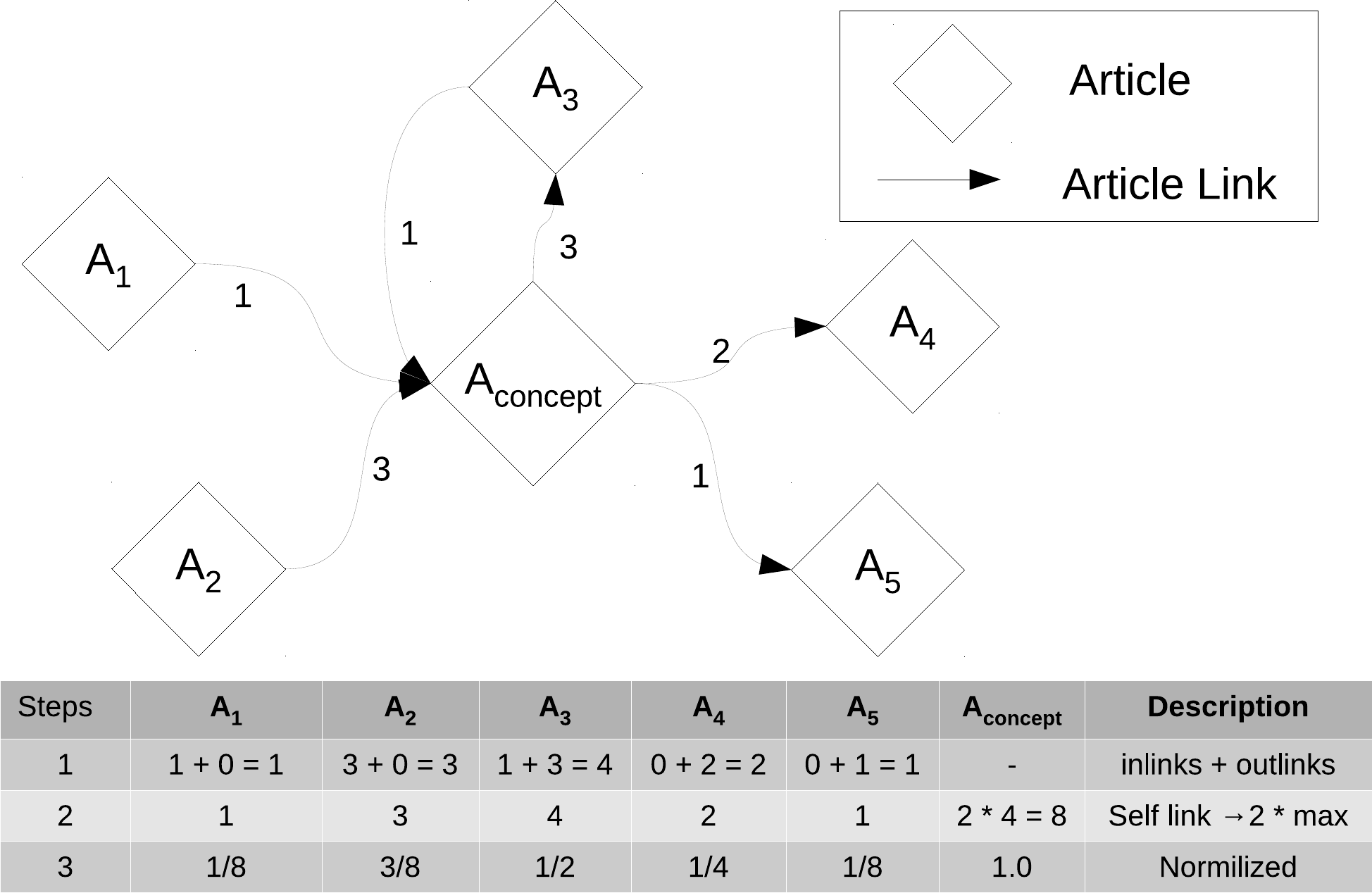}
\end{center}
\caption{An example of the assignment of the normalized $article_{score}$ for the concept article $A_{concept}$, based on inlink and outlink structure.}
\label{fig:eve-article}
\end{figure}

Fig. \ref{fig:eve-article} shows an example of the strategy. In the first step, all  inlinks and outlinks are counted for the other non-concept articles (\eg $A_{concept}$ has 3 inlinks and 1 outlink from $A_3$). In the next step, the self-link score is computed as twice the maximum of sum of inlinks and outlinks from all other articles (which is 8 in this case). In the final step, normalization\footnote{In case of best match strategy, where more than one article is mapped to a concept i.e., $A_{concept1}, A_{concept2}, ...$  the score computed is further scaled by the relevance score of the each article for the top-k articles, then reduced by the vector addition, and normalized again.} of the scores takes place, dividing by the maximum score (which is 8 in this case). Articles having no links to or from $A_{concept}$ receive a score of 0. Given the sparsity of the Wikipedia link graph, the article-based dimensions are also naturally sparse.

\subsubsection{Vector dimensions related to Wikipedia categories}
\label{sec:eve-category}
\begin{figure}
\begin{center}
\includegraphics[scale=0.42]{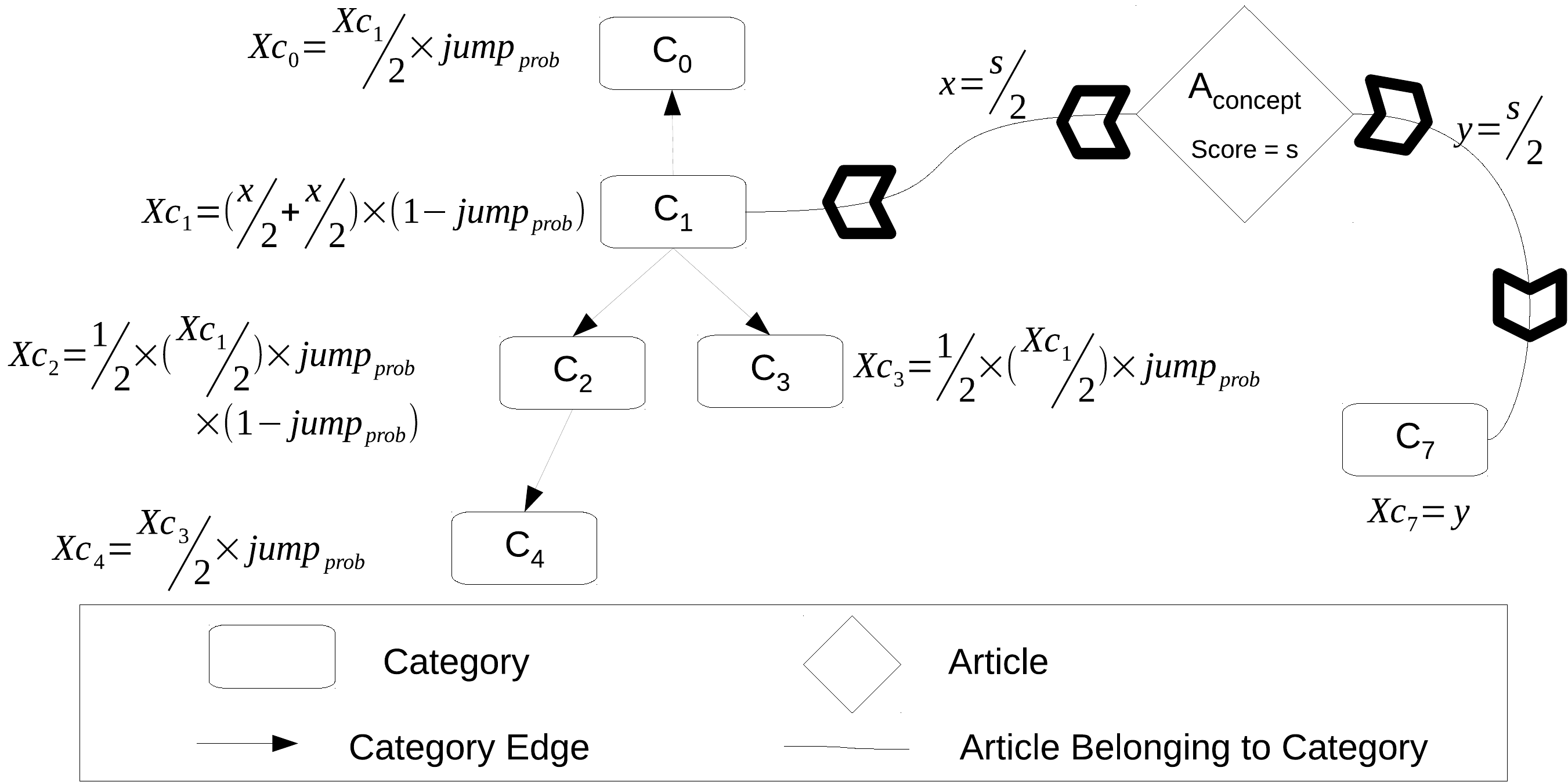}
\end{center}
\caption{Assignment of scores for the category dimensions, from the mapped article to its related categories.}
\label{fig:eve-category}
\end{figure}

Next we define the method for generating vector dimensions corresponding to all Wikipedia categories which are related to the concept article. The strategy to assign a score to the related Wikipedia categories proceeds as follows:
\begin{enumerate}
\item Start by propagating the score uniformly to the categories to which the concept article belongs to (see Fig. \ref{fig:WCG}).
\item A portion of the score is further propagated by the probability of jumping from a category to the categories in the neighborhood.
\item Score propagation continues until a certain hop count is reached (\ie a threshold value $category_{depth}$), or there are no further categories in the neighborhood. 
\end{enumerate}
Fig. \ref{fig:eve-category} illustrates the process, where the concept article $A_{concept}$ has a score \textit{s}, which is 1\footnote{In case of the partial best match it is the relevance score returned by \textit{BM25} algorithm.} for an exact match. First, the score is uniformly propagated across the number of Wikipedia categories and their tree structure to which the article belongs to ($C_1$ and $C_7$ tree receive $s/2$ from $A_{concept}$). In the next step, the directly-related categories ($C_1$ and $C_7$) further propagate the score to their super and sub-categories, while retaining a portion of score. $C_1$ retains a portion by the factor $1-jump_{prob}$ of the score that it propagate to the super and sub-categories. Where $jump_{prob}$ is the probability of jumping from a category to either a connected super or sub-category. While $C_7$ retains the full score since there is no super or sub-category for further propagation. In step 3 and onwards, the score continues to propagate in a direction (to either a super or sub-category) until hop count $category_{depth}$ is reached, or until there is no further category to which score could propagate to. In Fig. \ref{fig:eve-category}, $C_0$ and $C_3$ are the cases where the score cannot propagate further, while $C_4$ is the stopping condition for the score to propagate when using a threshold $category_{depth}=2$.

\subsubsection{Overall vector dimensions}

Once the sets of dimensions for related Wikipedia articles and categories have been created, we construct an overall vector for the concept article as follows. Eq. \ref{eq:eve-vector} shows the vector representation of a concept, where \textit{norm} is a normalization function, $articles_{score}$ and  $categories_{score}$ are the two sets of dimensions, while $bias_{article}$ and $bias_{category}$ are the bias weights which control the importance of the associations with the Wikipedia articles and categories respectively. The bias weights can tuned to give more importance to either type of association. In Eq. \ref{eq:eve-vector-norm} we normalize the entire vector such that the sum of the scores of all dimension equates to 1, so that a unit length vector is obtained. 
\begin{align}
	&\begin{aligned}
		\mathllap{Vector(concept)} &= <norm(articles_{score}) * bias_{article},\\
    	&\qquad norm(categories_{score}) * bias_{category}> \label{eq:eve-vector}\\
    \end{aligned}\\
    &\begin{aligned}
    \nonumber
    \end{aligned}\\
    &\begin{aligned}
		\mathllap{Vector(concept)} &= norm(Vector(concept)) \label{eq:eve-vector-norm} \\
    \end{aligned}
\end{align}
%
The process above is repeated for each word or concept in the input dataset to generate a set of vectors, representing an embedding of the data. In this embedding, each vector dimension is labeled with a tag which corresponds to either a Wikipedia article name or a Wikipedia category name. Therefore, each dimension carries a direct human-interpretable meaning. As we see in the next section, these labeled dimensions prove useful for the generation of algorithmic explanations.

\section{Evaluation}
In this section we investigate the extent to which embeddings generated using the \textit{EVE} model are useful in three fundamental data mining tasks. Firstly, we describe a number of alternative baseline methods, along with the relevant parameter settings. Then we describe the dataset which is used for the evaluations, and finally we report the experimental results to showcase the effectiveness of the model. We also highlight the benefits of the explanations generated as part of this process.

\subsection{Baselines and Parameters}
We compare \textit{EVE} with three popular word embedding algorithms: \textit{Word2Vec}, \textit{FastText}, and \textit{GloVe}. For \textit{Word2Vec} and \textit{FastText}, we trained two well-known variants of each -- \ie the continuous bag of words model (CBOW) and the skip-gram model (SG).  For \textit{GloVe}, we trained the standard model. 

For each baseline, we use the default implementation parameter values (window\_size=5, vector\_dimensions=100), except for the minimum document frequency threshold, which is set to 1 to generate all word vectors, even for rare words. This enables direct comparisons to be made with \textit{EVE}. For \textit{EVE}, we use uniform bias weights (\ie $bias_{article}$=0.5, $bias_{category}$=0.5). This provides equal importance to both dimension types. The parameter $jump_{prob}$=0.5 was chosen arbitrarily, so as to retain half of the score by the category while the rest is propagated. 




\subsection{Dataset}
To evaluate the performance of the different models, we constructed a new dataset from the complete 2015 English-language Wikipedia dump, composed of seven different topical types, each containing at least five sub-topical categories. On average each sub-topical category contains a list of 20 items or concepts. The usefulness of the dataset lies in the fact that the organization, from topics to categories to items, is made on the bases of factual position. 

\begin{table}[!t]
\caption{Summary statistics of the dataset.}
\label{tab:stats-dataset}       
\begin{tabular}{lccl}
\hline\noalign{\smallskip}
Topical Type & Categories & Mean Items per & Example (Category: Items) \\
  &  & Category& \\
\noalign{\smallskip}\hline\noalign{\smallskip}
Animal class & 5 & 20 & Mammal: Baleen whale, Elephant, \\
 &  &  & Primate \\
Continent to country & 6 & 17  & Europe: Albania, Belgium, Bulgaria \\
Cuisine & 5 & 20  & Italian cuisine: Agnolotti, Pasta, Pizza \\
European cities & 5 & 20  & Germany: Berlin, Bielefeld, Bonn \\
Movie genres  & 5 & 20  & Science fiction film: RoboCop, \\
 &  &   & The Matrix, Westworld \\
Music genres & 5 & 20  & Grunge: Alice in Chains \\
 &  &  & Chris Cornell, Eddie Vedder \\
Nobel laureates & 5 & 20  &  Nobel laureates in Physics: \\
 & &  & Albert Einstein, Niels Bohr\\
\noalign{\smallskip}\hline
\end{tabular}
\end{table}

\begin{table}[!t]
\caption{Dataset topical types and corresponding sub-topical categories.}
\label{tab:topical-categories-dataset}       
\begin{tabular}{ll}
\hline\noalign{\smallskip}
Topical Type & Categories\\ 
\hline\noalign{\smallskip}
Animal classes & Mammal, Reptile, Bird, Amphibian, Fish\\ 
Continent to Country & Africa, Europe, Asia, South America, North America, Oceania\\ 
Cuisine & Italian cuisine, Mexican cuisine, Pakistani cuisine, \\
 & Swedish cuisine, Vietnamese cuisine\\ 
European cities & France, Germany, Great Britain, Italy, Spain\\ 
Movie genres & Animation, Crime film, Horror film, Science fiction film,\\
& Western (genre)\\ 
Music genres & Jazz, Classical music, Grunge, Hip hop music, Britpop\\ 
Nobel laureates & Nobel laureates in Chemistry, Nobel Memorial \\
 & Prize laureates in Economics, Nobel laureates in Literature,\\
 & Nobel Peace Prize laureates, Nobel laureates in Physics\\ 
\noalign{\smallskip}\hline
\end{tabular}
\end{table}

Table \ref{tab:stats-dataset} shows a statistical summary of the dataset. In this table, the column ``Example (Category, Items)'' shows an example of a category name in the ``Topical Type'', together with a subset of list of items belonging to that category. For instance, in the first row ``Topical Type'' is \textit{Animal class} and \textit{Mammal} is one of the category belonging to this type, while \textit{Baleen whale} is an item with in the category of \textit{Mammal}. Similarly there are other categories of the type \textit{Animal class} such as \textit{Reptile}. Table \ref{tab:topical-categories-dataset} shows the list of categories for each topical type. 

All embedding algorithms in our comparison were trained on this dataset. In case of baseline models, we use ``article labels'', ``article redirects'', ``category labels'', and ``long abstracts'', with each entry as a separate document. Note that, prior to training, we filter out four non-informative Wikipedia categories which can be viewed as being analogous to stopwords: \{``articles contain video clips'', ``hidden categories'', ``articles created via the article wizard'', ``unprintworthy redirects''\}. 

\subsection{Experiments}

To compare the \textit{EVE} model with the various baseline methods, we define three general purpose data mining tasks: intruder detection, ability to cluster, ability to sort relevant items first. In the following sections we define the tasks separately,  each accompanied by experimental results and explanations.

\subsubsection{Experiment 1: Intruder detection}

First we evaluate the performance of \textit{EVE} when attempting to detect an unrelated ``intruder'' item from a list of \textit{n} items, where the rest of the items in the list are semantically related to one another. The ground truth for the correct relations between articles are based on the ``topical types'' in the dataset.

\paragraph{Task definition:}

For a given ``topical type'', we randomly choose four items belonging to one category and one intruder item from a different category of the same ``topical type''. After repeating this process exhaustively for all combinations for all topical types, we generated 13,532,280 results for this task. Table \ref{tab:queries-intruder} shows the breakdown of the total number of queries for each of the ``topical types''.

\paragraph{Example of a query:} For the ``topical type'' \textit{European cities}, we randomly choose four related items from the ``category'' \textit{Great Britain} such as \textit{London, Birmingham, Manchester, Liverpool}, while we randomly choose an intruder item \textit{Berlin} from the ``category'' \textit{Germany}. Each of the models is presented with the five items, where the challenge is to identify \textit{Berlin} as the intruder -- the rest of the items are related to each other as they are cities in \textit{Great Britain}, while \textit{Berlin} is a city in \textit{Germany}.

\paragraph{Strategy:}
In order to discover the intruder item, we formulate the problem as a maximization of pairwise similarity across all items, the item receiving the least score is least similar to all other items, and thus identified as the intruder. Formally, for each model we compute
\begin{align}
	score(item_{(k)}) &= \sum_{i=1}^{5} similarity(item_{(k)}, item_{(i)}); i \neq k \label{eq:intruder-strategy}
\end{align}
where the $similarity$ function is \textit{cosine similarity} \citep{Manning:2008:IIR:1394399}, $k$ and $i$ are the item positions in the list of items, and $item_{(k)}$ and $item_{(i)}$ are the vectors returned by the model under consideration. 

\begin{table}[!t]
\caption{\textit{Intruder detection} task --- Statistics for the number of queries.}
\label{tab:queries-intruder}       
\begin{tabular}{ll}
\hline\noalign{\smallskip}
Topical Types & No. of Queries \\
\hline\noalign{\smallskip}
Animal class & 1,938,000 \\
Continent to country & 1,904,280 \\
Cuisine & 1,938,000 \\
European cities  & 1,938,000 \\
Movie genres & 1,938,000 \\
Music genres & 1,938,000 \\
Nobel laureates & 1,938,000 \\
\noalign{\smallskip}\hline
Total & 13,532,280 \\
\noalign{\smallskip}\hline
\end{tabular}
\end{table}

\paragraph{Results:}

To evaluate the effectiveness of the \textit{EVE} model against the baselines for this task, we use $accuracy$ \citep{Manning:2008:IIR:1394399} as the measure for finding the intruder item. \textit{Accuracy} is defined as the ratio of correct results (or correct number of intruder items) to the total number of results returned by the model:
\begin{align}
	accuracy = \dfrac{\mid Results_{Correct}\mid}{\mid Results_{Total}\mid} \label{eq:accuracy}
\end{align}
Table \ref{tab:res-intrusion} shows the experimental results for the six models in this task. From the table it is evident that the \textit{EVE} model significantly outperforms rest of the models overall. However, in the case of two ``topical types'', the FastText CBOW yields better results. To explain this, we next show explanations generated by the \textit{EVE} model while making decisions for the intruder detection task.

\begin{table}[!t]
\caption{\textit{Intruder detection} task --- Detection accuracy results.}
\label{tab:res-intrusion}       
\begin{tabular}{lllllll}
\hline\noalign{\smallskip}
 & EVE & Word2Vec & Word2Vec & FastText & FastText & GloVe \\
  &  & CBOW & SG & CBOW & SG &  \\
\noalign{\smallskip}\hline
Animal class & \textbf{0.77} & 0.39 & 0.42 & 0.36 & 0.43 & 0.31 \\
Continent to Country & 0.75 & 0.70 & 0.76 & \textbf{0.79} & \textbf{0.79} & 0.73 \\
Cuisine & \textbf{0.97} & 0.34 & 0.43 & 0.62 & 0.75 & 0.25 \\
European cities & 0.94 & 0.93 & 0.98 & 0.91 & \textbf{0.99} & 0.74 \\
Movie genres & \textbf{0.71} & 0.23 & 0.24 & 0.22 & 0.25 & 0.21 \\
Music genres & \textbf{0.87} & 0.56 & 0.59 & 0.50 & 0.57 & 0.38 \\
Nobel laureates & \textbf{0.91} & 0.28 & 0.28 & 0.23 & 0.27 & 0.24 \\
\noalign{\smallskip}\hline
Average & \textbf{0.85} & 0.50 & 0.52 & 0.52 & 0.58 & 0.41 \\
\noalign{\smallskip}\hline
\end{tabular}
\begin{tablenotes}
      \small
      \item Note: all p-values are \num{<e-157} for EVE with respect to all baselines
    \end{tablenotes}
\end{table}

\paragraph{Explanation from the \textit{EVE} model:}

Using the labeled dimensions in vectors produced by \textit{EVE}, we define the process to generate effective explanations for the \textit{intruder detection} task in Algorithm \ref{algo:exp-intrusion} as follows. The inputs to this algorithm are the vectors of items, and the intruder item identified by the \textit{EVE} model. In step 1, we calculate the mean vector of all the vectors. In step 2 and 3, we subtract the influence of intruder and mean of vectors from each other to obtain dominant vector spaces to represent detected coherent items and intruder item respectively. In step 4 and 5, we order the labeled dimensions by their informativeness (\ie the dimension with the highest score is the most informative dimension). Finally, we return a ranked list of informative vector dimensions for the both non-intruders and the intruder as an explanation for the output of the task.

    \begin{algorithm}[h]
        \caption{Explanation strategy for \textit{intruder detection} task}
        \label{algo:exp-intrusion}
        \begin{algorithmic}[1]
            \REQUIRE \textit{EVE} $\rightarrow vector_{space}$, $vector_{intruder}$
            \STATE $space_{mean}$ = $Mean(vector_{space})$
            \STATE $coherentSpace_{leftover}$ = $space_{mean}$ - $vector_{intruder}$
            \STATE $intruder_{leftover}$ = $vector_{intruder}$ - $space_{mean}$
            \STATE $coherentSpace_{info\_features}$ = $order\_by_{info\_features} (coherentSpace_{leftover})$
            \STATE $intruder_{info\_features}$ = $order\_by_{info\_features} (intruder_{leftover})$
            \RETURN $coherentSpace_{info\_features}$, $intruder_{info\_features}$
        \end{algorithmic}
    \end{algorithm}
    
\begin{table}[!t]
\caption{Sample explanation generated for the \textit{intruder detection task}, for the query: \{Hawk, Penguin, Gull, Parrot, Snake\}. Correct intruder detected: Snake. All top-9 features are Wikipedia categories.}
\label{tab:exp-intrusion-corr}       
\renewcommand{\arraystretch}{1}
\begin{tabular}{|l|l|}
\hline 
\textbf{Non-Intruder} & \textbf{Intruder}\\ 
\hline
falconiformes & turonian first appearances\\ 
\hline
\textbf{birds of prey} & \textbf{snakes}\\ 
\hline
\textbf{seabirds} & squamata\\ 
\hline
ypresian first appearances & \textbf{predators}\\ 
\hline
psittaciformes & lepidosaurs\\ 
\hline
\textbf{parrots} & predation\\ 
\hline
rupelian first appearances & \textbf{carnivorous animals}\\ 
\hline
\textbf{gulls} & \textbf{venomous snakes}\\ 
\hline
\textbf{bird families} & \textbf{snakes in art}\\ 
\hline
\end{tabular}
\end{table}

Table \ref{tab:exp-intrusion-corr} and  \ref{tab:exp-intrusion-incorr} show sample explanations generated by the \textit{EVE} model, where the model has detected a correct and incorrect intruder item respectively. 
In Table \ref{tab:exp-intrusion-corr}, the query has items selected from ``topical type'' \textit{animal class}, where four of the items belong to the ``category'' \textit{birds}, while the item `snake' belongs to the ``category'' \textit{reptile}. As can be seen from the table, the bold features in the non-intruder and intruder column obviously represent bird family and snake respectively, which is the correct inference. Furthermore, the non-bold features in the non-intruder and intruder columns represent deeper relevant relations which may require some domain expertise. For instance, \textit{falconiformes} are a family of 60+ species in the order of birds and \textit{turonian} is the evolutionary era of the specific genera.




\begin{table}[!t]
\caption{Sample explanation generated for the \textit{intruder detection task}, for the query: \{I Am Legend (film), Insidious (film), A Nightmare on Elm Street, Final Destination (film), Children of Men\}. Incorrect intruder detected: Final Destination (film). All top-9 features are Wikipedia categories.
}
\label{tab:exp-intrusion-incorr}       
\renewcommand{\arraystretch}{1}
\begin{tabular}{|l|l|}
\hline
\textbf{Non-Intruder} & \textbf{Intruder}\\ 
\hline
\textbf{english-language films} & studiocanal films\\ 
\hline
american independent films & splatter films\\ 
\hline
american horror films & final destination films\\ 
\hline
\textbf{universal pictures films} & \textbf{films shot in vancouver}\\ 
\hline
\textbf{post-apocalyptic films} & \textbf{films shot in toronto}\\ 
\hline
\textbf{films based on science fiction novels} & films shot in san francisco, california\\ 
\hline
\textbf{2000s science fiction films} & films set in new york\\ 
\hline
ghost films & films set in 1999\\ 
\hline
films shot in los angeles, california & film scores by shirley walker\\ 
\hline
\end{tabular}
\end{table}

In the example in Table \ref{tab:exp-intrusion-incorr}, the query has items selected from the ``topical type'' \textit{movie genres}, where four of the items belong to the ``category'' \textit{horror film}, while the intruder item `Children of Men' belongs to the ``category'' \textit{science fiction film}. In this example, \textit{EVE} identifies the wrong intruder item according to the ground truth, recommending instead the item `Final Destination (film)'. From the explanation in the table, it becomes clear why the model made this recommendation. 
We observe that the non-intruder items have a coherent relationship with `post-apocalyptic films' and `films based on science fiction novels' (both `I am Legend (film)' and `Children of Men' belong to these categories). Whereas `Final Destination (film)' was recommended by the model based on features relating to filming location. A key advantage of having an explanation from the model is that it allows us to understand why a mistake occurs and how we might improve the model. In this case, one way to make improvement might be to add a rule filtering Wikipedia categories relating to locations when consider movie genres.

\subsubsection{Experiment 2: Ability to cluster}

In this experiment, we evaluate the extent to which the distances computed on \textit{EVE} embeddings can help to group semantically-related items together, while keeping unrelated items apart. This is a fundamental requirement for distance-based methods for cluster analysis.

\paragraph{Task definition:}

For all items in a specific ``topical type'', we construct an embedding space without using information about the category to which the items belong. The purpose is then to measure the extent to which these items cluster together in the space relative to the ground truth categories. This is done by measuring distances in the space between items that should belong together (\ie intra-cluster distances) and items that should be kept apart (\ie inter-cluster distances), as determine by the categories. Since there are seven ``topical types'', there are also even queries in this task.

\paragraph{Example of a query:}
For the ``topical type'' \textit{Cuisine}, we are provided with a list of 100 items in total, where each of the five categories has 20 items. These correspond to cuisine items from five different countries. The idea is to measure the ability of each embedding model to cluster these 100 items back into five categories. 

\paragraph{Strategy:}
To formally measure the ability of a model to cluster items, we conduct a two-step strategy as follows:
\begin{enumerate}
\item Calculate a pairwise similarity matrix between all items of a given ``topical type''. The similarity function that we use for this task is the \textit{cosine similarity}. 
\item Transform the similarity matrix to a distance matrix\footnote{by simply, 1 - normalized similarity score over each dimension} which is used to measure inter and intra-cluster distances relative to the ground truth categories.
\end{enumerate}

\paragraph{Results:}
To evaluate the ability to cluster, there are typically two objectives: within-cluster cohesion and between-cluster separation. To this end, we use three well-known cluster validity measures in this task. Firstly, the \emph{within-cluster distance} \citep{everitt2001cluster} is the total of the squared distances between each item $x_i$ and the centroid vector $\mu_c$ of the cluster $C_c$ to which it has been assigned:
\begin{align}
within = \sum_{c=1}^{k} \sum_{x_i \in C_c} d(x_i,\mu_c)^2
\end{align}
Typically this value is normalized with respect to the number of clusters $k$.
The higher the score, the more coherent the clusters. 
Secondly, the \emph{between-cluster distance} is the total of the squares of the distances between the each cluster centroid and the centroid of the entire dataset, denoted $\hat{\mu}$:

\begin{table}[!t]
\caption{\textit{Ability to cluster} task --- Mean within-cluster distance scores.}
\label{tab:mean-within-cluster}       
\begin{tabular}{lllllll}
\hline\noalign{\smallskip}
 & EVE & Word2Vec & Word2Vec & FastText & FastText & GloVe \\
  &  & CBOW & SG & CBOW & SG &  \\
\noalign{\smallskip}\hline
Animal class & \textbf{2.00} & 13.03 & 6.23 & 10.31 & 7.71 & 12.20\\ 
Continent to country & 2.34 & 2.63 & \textbf{2.25} & 2.83 & 2.56 & 2.60\\ 
Cuisine & \textbf{2.92} & 17.31 & 8.88 & 9.74 & 6.25 & 12.36\\ 
European cities & \textbf{3.13} & 7.72 & 5.46 & 8.30 & 5.75 & 6.86\\ 
Movie genres & 6.92 & 11.98 & \textbf{6.04} & 9.81 & 5.61 & 17.96\\ 
Music genres & \textbf{1.90} & 8.25 & 5.25 & 6.72 & 5.77 & 7.72\\ 
Nobel laureates & \textbf{2.88} & 14.56 & 8.99 & 12.40 & 10.59 & 15.13\\ 
\noalign{\smallskip}\hline
Average & \textbf{3.16} & 10.78 & 6.16 & 8.59 & 6.32 & 10.69\\ 
\noalign{\smallskip}\hline
\end{tabular}
\end{table}

\begin{table}[!t]
\caption{\textit{Ability to cluster} task --- Mean between-cluster distance scores.}
\label{tab:mean-between-cluster}       
\begin{tabular}{lllllll}
\hline\noalign{\smallskip}
 & EVE & Word2Vec & Word2Vec & FastText & FastText & GloVe \\
  &  & CBOW & SG & CBOW & SG &  \\
\noalign{\smallskip}\hline
Animal class & 0.47 & \textbf{1.30} & 0.74 & 1.14 & 1.13 & 0.46\\ 
Continent to country & 3.33 & 3.86 & 1.78 & \textbf{4.08} & 2.83 & 1.63\\ 
Cuisine & 8.18 & 2.12 & 2.12 & \textbf{14.52} & 10.80 & 0.88\\ 
European cities & 2.39 & \textbf{17.14} & 7.45 & 13.24 & 10.86 & 3.84\\ 
Movie genres & \textbf{1.58} & 0.40 & 0.18 & 0.41 & 0.18 & 0.48\\ 
Music genres & 2.23 & \textbf{2.79} & 1.60 & 1.16 & 0.18 & 1.68\\ 
Nobel laureates & \textbf{1.95} & 0.79 & 0.39 & 0.56 & 1.38 & 0.20\\ 
\noalign{\smallskip}\hline
Average & 2.88 & 4.06 & 2.04 & \textbf{5.02} & 3.96 & 1.31\\ 
\noalign{\smallskip}\hline
\end{tabular}
\end{table}

\begin{table}[!t]
\caption{\textit{Ability to cluster} task --- Overall CH-Index validation scores.}
\label{tab:ch-index-cluster}       
\begin{tabular}{lllllll}
\hline\noalign{\smallskip}
 & EVE & Word2Vec & Word2Vec & FastText & FastText & GloVe \\
  &  & CBOW & SG & CBOW & SG &  \\
\noalign{\smallskip}\hline
Animal class & \textbf{7.64} & 5.98 & 4.09 & 3.91 & 4.44 & 5.46\\ 
Continent to country & \textbf{15.83} & 11.84 & 8.19 & 13.69 & 12.29 & 7.52\\ 
Cuisine & \textbf{54.18} & 2.38 & 3.51 & 14.25 & 16.00 & 2.23\\ 
European cities & 29.08 & \textbf{48.57} & 28.98 & 33.73 & 41.88 & 15.53\\ 
Movie genres & \textbf{12.45} & 1.36 & 1.43 & 1.51 & 1.87 & 1.27\\ 
Music genres & \textbf{25.04} & 18.01 & 14.80 & 13.06 & 12.93 & 6.09\\ 
Nobel laureates & \textbf{21.85} & 3.58 & 3.34 & 1.73 & 3.16 & 2.91\\ 
\noalign{\smallskip}\hline
Average & \textbf{23.72} & 13.10 & 9.19 & 11.70 & 13.22 & 5.86\\ 
\noalign{\smallskip}\hline
\end{tabular}
\end{table}

\begin{align}
between = \sum_{c=1}^{k} \left| C_c \right| \; d(\mu_c,\hat{\mu})^2 
\;\;\textrm{ where }\;\;
\hat{\mu} = \frac{1}{n} \sum_{i=1}^{n} x_i
\end{align}
This value is also normalized with respect to the number of clusters $k$.
The lower the score, the more well-separated the clusters. Finally, the two above objectives are combined via the \textit{CH-Index} \citep{calinski1974dendrite}, using the ratio:
\begin{align}
CH = \frac{ between / (k-1) }{ within / (n-k) }
\label{eqn:int:ch}
\end{align}
The higher the value of this measure, the better the overall clustering.


From Table \ref{tab:mean-within-cluster}, we can see that \textit{EVE} generally performs better than rest of the embedding methods for the \textit{within-cluster} measure. In Table \ref{tab:mean-between-cluster}, for the \textit{between-cluster} measure, \textit{EVE} is outperformed by \textit{FastText CBOW}, \textit{Word2Vec CBOW}, and \textit{FastText SG} mainly due to the ``topical type'' \textit{Cuisine} and \textit{European cities} where \textit{EVE} does not perform well. Finally, in Table \ref{tab:ch-index-cluster} where the combined aim of clustering is captured through the \textit{CH-Index}, \textit{EVE} outperforms the rest of the methods, except in the case of the ``topical type'' \textit{European cities}.

\paragraph{Explanation from the \textit{EVE} model:}
Using labeled dimensions from the \textit{EVE} model, we define a similar strategy for explanation as used in the previous task. However, now instead of discovering an intruder item, the goal is to define categories from items and to define the overall space. Algorithm \ref{algo:exp-cluster} shows the strategy which requires three inputs: the $vector{space}$ representing the entire embedding; the list of categories $categories$; the $categories\_vector_{space}$ which is the vector space of items belonging to each category. 
In step 1, we calculate the mean vector representing for the entire space. In step 2, we order the labeled dimensions of the mean vector by the informativeness. In steps 3--6 we iterate over the list of categories (of a ``topical type'' such as \textit{Cuisine}) and calculate mean vector for each category's vector space, which is followed by the ordering of dimensions of the mean vector of category vector space by the informativeness. Finally, we return the most informative features of the entire space and of each category's vector space.

    \begin{algorithm}[h]
        \caption{Explanation strategy for the \textit{ability to cluster} task.}
        \label{algo:exp-cluster}
        \begin{algorithmic}[1]
            \REQUIRE \textit{EVE} $\rightarrow vector_{space}$, $categories$, $categories\_vector_{space}$
            \STATE $space_{mean}$ = $Mean(vector_{space})$
            \STATE $space_{info\_features}$ = $order\_by_{info\_features} (space_{mean})$
            \FOR{$category \in categories$}
              \STATE $category_{mean}$ = $Mean(categories_\_vector_{space}[category])$
              \STATE $categories_{info\_features}[category]$ = $order\_by_{info\_features} (category_{mean})$
            \ENDFOR
            \RETURN $space_{info\_features}$, $categories_{info\_features}$
        \end{algorithmic}
    \end{algorithm}

\begin{table}
\caption{Sample explanation generated for the \textit{ability to cluster} task, for the query:\{items of ``topical type'' \textit{Cuisine}\}. All top-6 features are Wikipedia categories, except for those beginning with `$\alpha$:' which correspond to Wikipedia articles.}

\label{tab:exp-cluster-best}       
\renewcommand{\arraystretch}{1}
\begin{tabular}{|l|l|l|l|l|l|}
\hline
\textbf{Overall} & \textbf{Italian} & \textbf{Mexican} & \textbf{Pakistani} & \textbf{Swedish} & \textbf{Vietnamese}\\ 
\textbf{space} & \textbf{category} & \textbf{category} & \textbf{category} & \textbf{category} & \textbf{category}\\ 
\hline
\textbf{vietnamese}  & \textbf{italian}  & \textbf{mexican}  & \textbf{pakistani}  & \textbf{swedish} & \textbf{vietnamese} \\ 
 \textbf{cuisine} & \textbf{cuisine} &  \textbf{cuisine} &  \textbf{cuisine} &  \textbf{cuisine} &  \textbf{cuisine}\\
 \hline
 \textbf{swedish} & cuisine & tortilla-  & indian  & finnish  &  vietnamese\\ 
 \textbf{cuisine} & of lombardy & based & cuisine &  cuisine & words and \\ 
  &  & dishes &  &  & phrases\\ 
  \hline
  \textbf{mexican}  & types of  & cuisine of & indian  & \textbf{$\alpha$:swedish}  & \textbf{$\alpha$:} \\ 
  \textbf{cuisine} &  pasta & the south- &  desserts & \textbf{cuisine} &  \textbf{vietnamese} \\ 
  &   & western &   &   & \textbf{cuisine}\\ 
  &   & united states &   &   & \\ 
  \hline
 \textbf{italian}  & pasta & cuisine of & \textbf{pakistani}  & desserts & $\alpha$:vietnam\\ 
 \textbf{cuisine} &  & the western  &  \textbf{breads} &  & \\ 
  &  &  united states &   &  & \\ 
  \hline
 dumplings & dumplings & \textbf{$\alpha$:list of}  & iranian  & $\alpha$:sweden & $\alpha$:gÃÂ nÃÂÃÂ°ÃÂ¡ÃÂ»ÃÂng sÃÂ¡ÃÂºÃÂ£\\ 
 &  & \textbf{mexican}  &  breads &  & \\ 
 &  &  \textbf{dishes} &   &  & \\ 
 \hline
 \textbf{pakistani}  & \textbf{italian-}  & maize  & \textbf{pakistani}  & potato  & $\alpha$:thit kho \\ 
 \textbf{cuisine} &  \textbf{american}  &  dishes &  \textbf{meat} &  dishes & tau\\ 
 &   \textbf{cuisine} &   & \textbf{dishes} &   & \\ 
\hline
\end{tabular}
\end{table}

Tables \ref{tab:exp-cluster-best} and  \ref{tab:exp-cluster-worse} show the explanations generated by the \textit{EVE} model, in the cases where the model performed best and worse against baselines respectively. 
In Table \ref{tab:exp-cluster-best}, the query is the list of items from ``topical type'' \textit{cuisine}. As can be seen from the bold entries in the table, the explanation conveys the main idea about both the overall space and the individual categories. For example, in the overall space, we can see the cuisines by different nationalities, and likewise we can see the name of nationality from which the cuisine is originated from (\eg \textit{Italian cuisine} for the ``Italian category'' and \textit{Pakistani breads} for the ``Pakistani category''). As for the non-bold entries, we can also observe relevant features but at a deeper semantic level. For example, \textit{cuisine of Lombardy} in ``Italian category'' where Lombardy is a region in Italy, and likewise \textit{tortilla-based dishes} in the \textit{Mexican category} where tortilla is a primary ingredient in Mexican cuisine.

\begin{table}
\caption{Sample explanation for the \textit{ability to cluster} task, for the query: \{items of ``topical type'' \textit{European cities}\}. All top-6 features are Wikipedia categories.}

\label{tab:exp-cluster-worse}       
\renewcommand{\arraystretch}{1}
\begin{tabular}{|l|l|l|l|l|l|}
\hline
\textbf{Overall}  & \textbf{France}  & \textbf{Great }  & \textbf{Germany}  & \textbf{Italy}  & \textbf{Spain} \\ 
\textbf{space} & \textbf{category} & \textbf{Britain} & \textbf{category} & \textbf{category} & \textbf{category}\\ 
&  &  \textbf{category} &   &  & \\ 
\hline
prefectures   & prefectures   & articles & university & \textbf{world}  & university \\ 
in france & in france & including  & towns in  & \textbf{heritage}  & towns in \\ 
 &  & recorded  & germany &  \textbf{sites in}  & spain\\ 
  &  &  pronuncia- &  &   \textbf{italy}  & \\ 
 &   &  tions (uk &  &     & \\ 
 &   &  english) &  &     & \\ 
 \hline
university  & port cities & county towns & \textbf{members }  & mediterra- & populated \\ 
towns in & and towns & in england & \textbf{of the}  & nean port & coastal \\ 
germany & on the fren-& &  \textbf{hanseatic} & cities and & places in\\ 
 & ch atlantic & & \textbf{league} & towns in &  spain\\ 
 & coast & &   & italy &  \\ 
 \hline
\textbf{members}   & cities in & metropolitan & german  & populated & \textbf{roman}  \\ 
\textbf{of the} & france &  boroughs & state & coastal  & \textbf{sites in}\\ 
\textbf{hanseatic}  &  &   & capitals & places in & \textbf{spain} \\ 
\textbf{league} &  &   &  & italy &  \\ 
 \hline
articles & subpre- & university  & cities in  & cities and  & port cities \\ 
including  & fectures & towns in the &  north rhine- & towns in & and towns\\
recorded & in france & united  & westphalia & emilia & on the \\
pronuncia-&  &  kingdom &  & romagna & spanish \\
tions (uk  &  &   &  &  &  atlantic coast\\
english) &  &   &  &  &  coast\\
\hline
\textbf{capitals }& \textbf{world}  & populated  & rhine  & former  & tourism \\ 
\textbf{of former}  & \textbf{heritage}  &  places  &  province &  capitals & in spain\\ 
 \textbf{nations} & \textbf{sites in}  &  established  &   &   of italy & \\ 
  & \textbf{france} &   in the 1st &   &   & \\ 
  & &  century &   &   & \\ 
\hline
german  & communes  & \textbf{fortified}  & populated  & \textbf{capitals}  & mediterranean \\ 
state  &  of nord  &  \textbf{settlements} &  places on  & \textbf{of former}  & port cities \\ 
 capitals &  (french &   & the rhine &  \textbf{nations} & and towns in \\ 
  &   department) &   & &   &  spain\\ 
\hline
\end{tabular}
\end{table}

In Table \ref{tab:exp-cluster-worse}, the query is the list of items from ``topical type'' \textit{European cities} and this is the example where \textit{EVE} model performs worse. However, the explanation allows us to understand why this is the case. As can been from the explanation table, the bold features show historic relationships across different countries, such as ``capitals of former nations'', ``fortified settlements'', and ``Roman sites in Spain''. Similarly, it can also be observed in non-bold features such as ``former capital of Italy''. Based on this explanation, we could potentially decide to apply a rule that would exclude any historical articles or categories when generating the embedding for this type of task in future.

\begin{figure}%
    \centering
    \subfloat[\textit{EVE} model]{{\includegraphics[trim={4.5cm 4.2cm 2.2cm 3.2cm},clip, width=5.2cm]{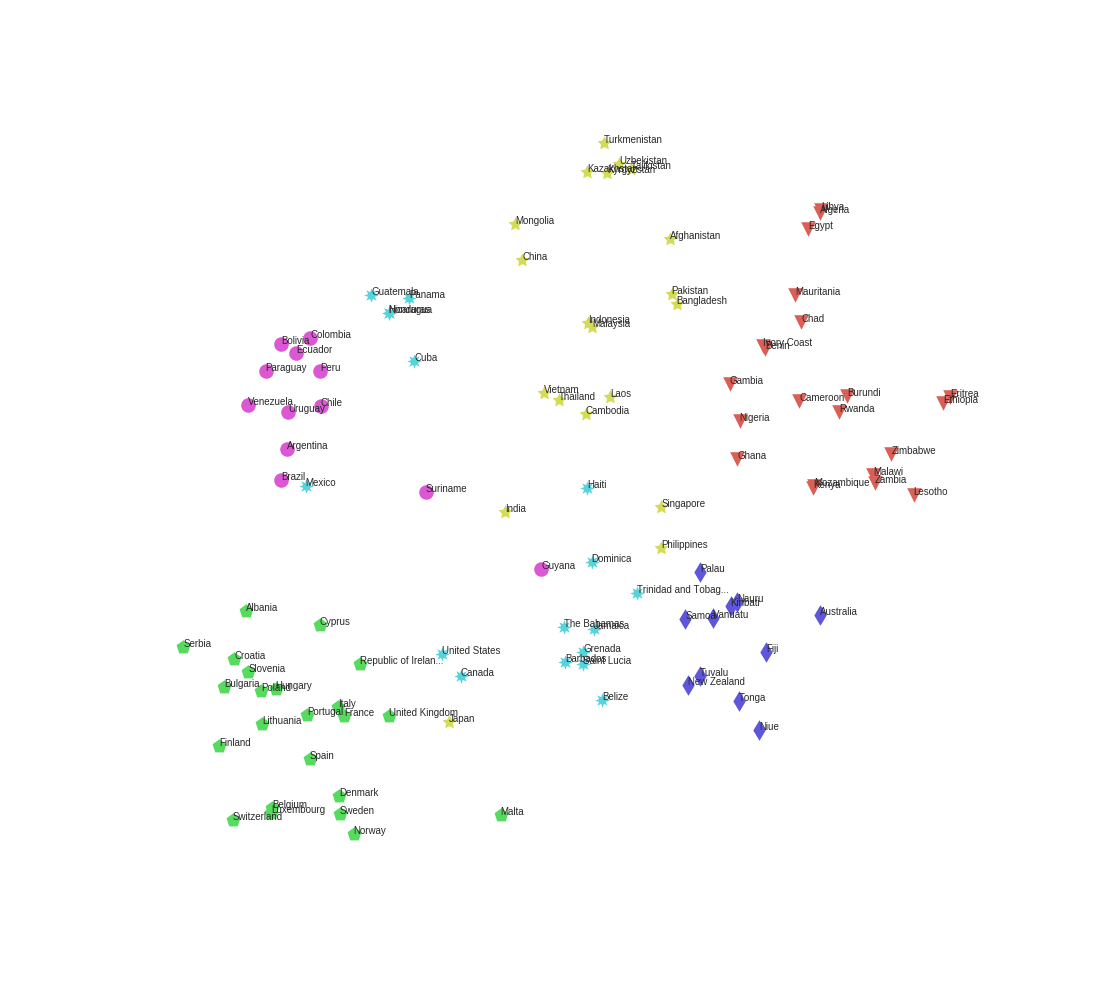} }}%
    \qquad
    \subfloat[\textit{GloVe} model]{{\includegraphics[trim={4.5cm 4.2cm 2.2cm 3.2cm},clip, width=5.2cm]{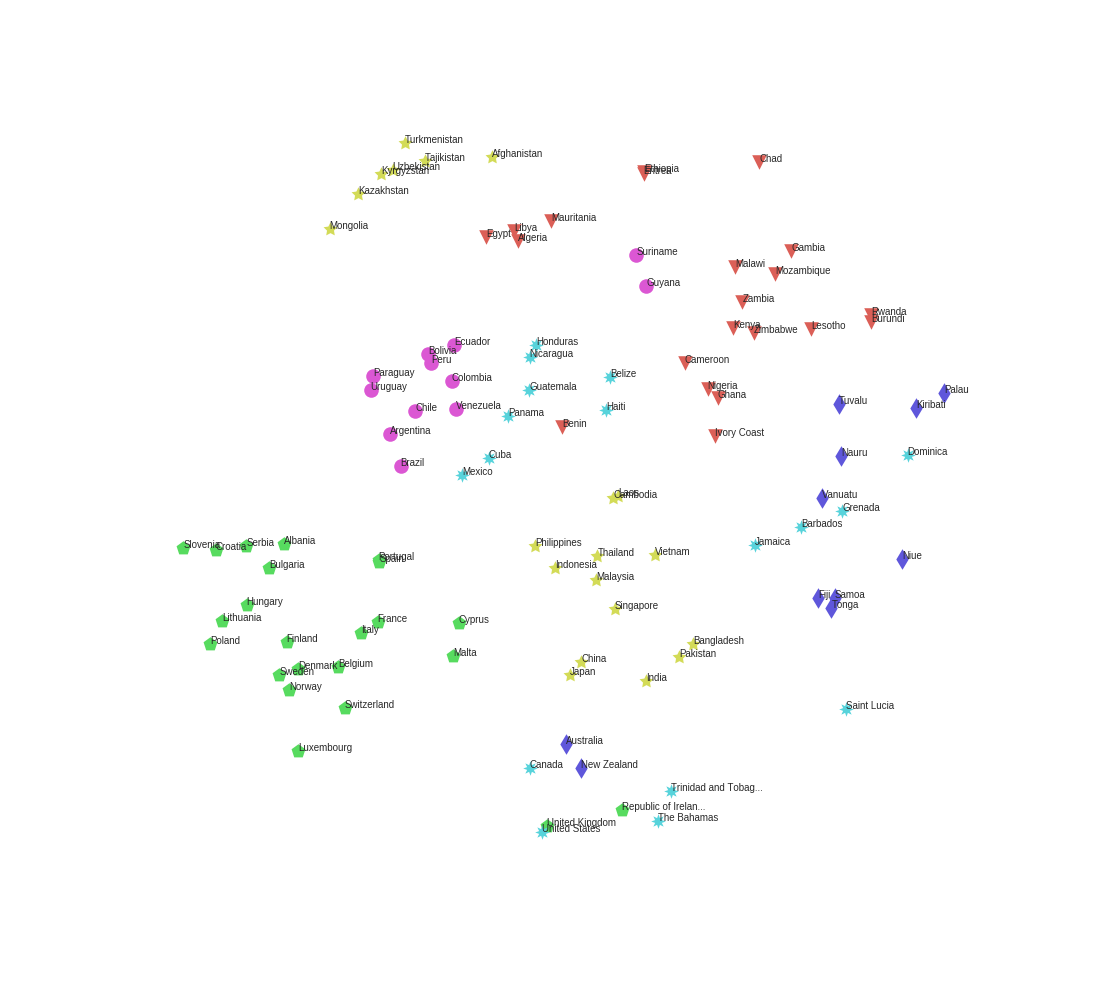} }}%
    \caption{Visualizations of model embeddings generated for \textit{the ability to cluster} task, for the query: \{items of ``topical type'' \textit{Country to Continent}\}. Colors and shapes indicate items belonging to different ground truth categories.}%
    \label{fig:cluster-visual}%
\end{figure}

\paragraph{Visualization:}
Since scatter plots are often used to represent the output of a cluster analysis process, we generate a visualization of all embeddings using T-SNE \citep{maaten2008tsne}, which is a tool to visually represent high-dimensional data by reducing it to 2--3 dimensions for presentation.\footnote{The full set of experimental visualizations is available at \url{http://mlg.ucd.ie/eve/}}. For the interest of reader, Fig. \ref{fig:cluster-visual} shows a visualization generated using \textit{EVE} and \textit{GloVe} when the list of items are selected from the ``topical type'' \textit{country to continent}. As can be seen from the plot, the ground truth categories exhibit better clustering behavior when using the space from the \textit{EVE} model, when compared to the \textit{Glove} model. This is also reflected in the corresponding scores in Tables \ref{tab:mean-within-cluster}, \ref{tab:mean-between-cluster}, and \ref{tab:ch-index-cluster}.

\subsubsection{Experiment 3: Sorting relevant items first}

\paragraph{Task definition:}
The objective of this task is to rank a list of items based on their relevance to a given query item. According to the ground truth associated with our dataset, items which belong to the same `category' of ``topical type'' as the query should be ranked above items which do not belong that `category' (\ie they are irrelevant to the query). In this task the total number of queries is equal to the total number of categories in the dataset -- \ie 36 (see table \ref{tab:stats-dataset}).

\paragraph{Example of a query:}
Unlike the previous tasks, here `category' is used as a query in this task. For example, for the `category' \textit{Nobel laureates in Physics}, the task is to sort all items from ``topical type'' \textit{Nobel laureates} such that the list of items from `category' \textit{Nobel laureates in Physics} are ranked ahead of the rest of the items. Thus, Niels Bohr, who is a laureate in Physics, should appear near the top of the ranking, unlike Elihu Root, who is a prize winner in Peace.

\paragraph{Strategy:}
In order to sort items relevant to a category, we define a simple two-step strategy as follows:
\begin{enumerate}
\item Calculate the \textit{cosine similarity} between all items and a category belonging to ``topical type'' in the model space.
\item Sort the list of items in descending order according to their similarity scores with the category.
\end{enumerate}
Based on this strategy, a successful model should rank items with the same `category' before irrelevant items.

\paragraph{Results:}
We use precision-at-$k$ ($P@k$) and average precision ($AP$) \citep{Manning:2008:IIR:1394399} as the measures to evaluate the effectiveness of the sorting ability of each embedding model with respect to relevance of items to a category. $P@k$ captures how many relevant items are calculated at a certain rank (or in the $top-k$ results), while $AP$ captures how early a relevant item is retrieved on average. It may happen that two models have the same value of $P@k$, while one of the models retrieves relevant items in an earlier order of rank, thus achieving a higher $AP$ value. $P@k$ is defined as the ratio of relevant items retrieved in the $top-k$ retrieved items, whereas $AP$ is the average of $P@k$ values computed after each relevant item is retrieved. Equations \ref{eq:p-k} and \ref{eq:ap} show the formal definitions of both measures.
\begin{align}
	P@k = \dfrac{\mid Items_{Relevant}\mid}{\mid Items_{Top\mbox{-}k}\mid} \label{eq:p-k}
\end{align}

\begin{align}
	&\begin{aligned}
      AP = \dfrac{1}{\mid Items_{Relevant} \mid}\sum_{k=1}^{\mid Items \mid} P@k \cdot rel(k) \label{eq:ap} \\
    \end{aligned}\\
    &\begin{aligned}
	\nonumber
      \text{where }rel(k) = 
          &\left\{
          \begin{array}{ll}
            1, \text{ if \textit{item(k)} is relevant}  \\ 
            0, \text{ otherwise} \        \end{array}
        \right.
  \end{aligned}
\end{align}

Tables \ref{tab:p-k-ranking-cluster} and \ref{tab:ap-ranking-cluster} show the experimental results of the \textit{sorting relevant items first} task. We choose $P@20$ ($k=20$), since on average there are 20 items in each category in the dataset. As can be seen from tables, the \textit{EVE} model generally outperforms the rest of models, except for the ``topical type'' \textit{European cities} where it gets outperformed by a factor of 1.05 and 1.09 times in terms of $P@k$ and $AP$ respectively, while in all other cases \textit{EVE} outperforms other algorithm by at least 1.51 and 1.37 times in terms of $P@k$ and $AP$ respectively. On average, the \textit{EVE} model outperforms the second best algorithm by a factor of 1.8 and 1.67 times in terms of $P@k$ and $AP$ respectively. In the next section, we show the corresponding explanations generated by the \textit{EVE} model for this task.

\begin{table}[!t]
\caption{\textit{Sorting relevant items first} task -- Precision ($P@20$) scores.}
\label{tab:p-k-ranking-cluster}       
\begin{tabular}{lllllll}
\hline\noalign{\smallskip}
 & EVE & Word2Vec & Word2Vec & FastText & FastText & GloVe 
 \\
  &  & CBOW & SG & CBOW & SG &  \\
\noalign{\smallskip}\hline
Animal class & \textbf{0.72} & 0.34 & 0.38 & 0.41 & 0.47 & 0.22\\ 
Continent to country & \textbf{0.95} & 0.54 & 0.51 & 0.63 & 0.59 & 0.31\\ 
Cuisine & \textbf{0.97} & 0.36 & 0.49 & 0.54 & 0.54 & 0.24\\ 
European cities & 0.91 & 0.85 & 0.91 & 0.86 & \textbf{0.96} & 0.61\\ 
Movie genres & \textbf{0.87} & 0.30 & 0.31 & 0.24 & 0.29 & 0.24\\ 
Music genres & \textbf{0.90} & 0.33 & 0.30 & 0.28 & 0.37 & 0.21\\ 
Nobel laureates & \textbf{0.99} & 0.27 & 0.22 & 0.20 & 0.25 & 0.20\\ 
\noalign{\smallskip}\hline
Average & \textbf{0.90} & 0.43 & 0.45 & 0.45 & 0.50 & 0.29\\ 
\noalign{\smallskip}\hline
\end{tabular}
\end{table}

\begin{table}[!t]
\caption{\textit{Sorting relevant items first} task -- Average Precision (AP) scores.}
\label{tab:ap-ranking-cluster}       
\begin{tabular}{lllllll}
\hline\noalign{\smallskip}
 & EVE & Word2Vec & Word2Vec & FastText & FastText & GloVe \\
  &  & CBOW & SG & CBOW & SG &  \\
\noalign{\smallskip}\hline
Animal class & \textbf{0.72} & 0.38 & 0.42 & 0.45 & 0.52 & 0.27\\ 
Continent to country & \textbf{0.92} & 0.55 & 0.54 & 0.65 & 0.67 & 0.33\\ 
Cuisine & \textbf{0.99} & 0.39 & 0.58 & 0.59 & 0.59 & 0.27\\ 
European cities & 0.91 & 0.91 & 0.97 & 0.93 & \textbf{0.99} & 0.65\\ 
Movie genres & \textbf{0.88} & 0.32 & 0.35 & 0.29 & 0.34 & 0.29\\ 
Music genres & \textbf{0.91} & 0.35 & 0.34 & 0.33 & 0.40 & 0.29\\ 
Nobel laureates & \textbf{1.00} & 0.26 & 0.26 & 0.24 & 0.29 & 0.24\\ 
\noalign{\smallskip}\hline
Average & \textbf{0.90} & 0.45 & 0.49 & 0.49 & 0.54 & 0.33\\ 
\noalign{\smallskip}\hline
\end{tabular}
\end{table}

\paragraph{Explanation from the \textit{EVE} model:}
Using the labeled dimensions provided by the \textit{EVE} model, we define a strategy for generating explanations for the \textit{sorting relevant items first} task in Algorithm \ref{algo:exp-sorting}. The strategy requires three inputs. The first is the $vector_{space}$ which is composed of category vector and item vectors. The second input is the $Sim_{wrt\_category}$ which is a column matrix, composed of similarity score between the category vector with itself and item vectors. In this matrix the first entry is 1.0 because of the self similarity of the category vector. The final input is a list of items $items$.
In the step 1 and 2, a weighted mean vector of space is calculated, where the weights are the similarity scores between the vectors in the space and the category vector. In steps 3--6, we iterate over the list of items and calculate the product between the weighted mean vector of the space and the item vector. After taking the product, we order the dimensions by the informativeness. Finally, we return the ranked list of informative features for each item.

    \begin{algorithm}[h]
        \caption{Explanation strategy for \textit{sorting relevant items first} task}
        \label{algo:exp-sorting}
        \begin{algorithmic}[1]
            \REQUIRE \textit{EVE} $\rightarrow vector_{space}, Sim_{wrt\_category}$, $items$
            \STATE $BiasedSpace$ = $vector_{space} \times SimilarityMatrix$
            \STATE $BiasedSpace_{mean}$ = $Mean(BiasedSpace)$            
            \FOR{$item \in items$}
              \STATE $item_{projection}$ = $BiasedSpace_{mean} \times vector_{space}[item]^T$
              \STATE $items_{info\_features}[item]$ = $order\_by_{info\_features} (item_{projection})$
            \ENDFOR
            \RETURN $items_{info\_features}$
            
        \end{algorithmic}
    \end{algorithm}

Tables \ref{tab:exp-sorting-nobel} and  \ref{tab:exp-sorting-music} show sample explanations generated by the \textit{EVE} model. For illustration purposes we select the ``topical types'' \textit{Nobel laureates} and \textit{Music genres} for explanations, as these are the only remaining ``topical types'' which we have not looked at so far in the other tasks.
\begin{table}
\caption{Sample explanation for the \textit{sorting relevant items first} task, for the query: \{Nobel laureates in Chemistry\}. All top-6 features are Wikipedia categories.
}
\label{tab:exp-sorting-nobel}       
\begin{tabular}{|l|l|}
\hline
\textbf{Kurt Alder (Chemistry)} & \textbf{Linus Pauling (Peace)}\\  
\textbf{First correct found at \#1} & \textbf{First incorrect found at \#20}\\  
\hline
\textbf{nobel laureates in chemistry} & \textbf{nobel laureates in chemistry}\\ 
\hline
german nobel laureates & Guggenheim fellows\\
\hline
\textbf{organic chemists} & american nobel laureates\\ 
\hline
university of kiel faculty & national medal of science laureates\\  
\hline
university of kiel alumni & american physical chemists\\ 
\hline
university of cologne faculty & american people of scottish descent\\ 
\hline
\end{tabular}
\end{table}

In Table \ref{tab:exp-sorting-nobel}, the query is `category' \textit{Nobel laureates in Chemistry} from the ``topical type'' \textit{nobel laureates}. We show the informative features for two cases -- the first correct result which appears at rank 1 in the sorted lists produced by \textit{EVE}, and the first incorrect result which appears at rank 20. The bold features indicates that both individuals are Nobel laureates in Chemistry. However, Linus Pauling also appears to be associated with the Peace category. This reflects that fact that, in fact, Linus Pauling is a two time Nobel laureate in two different categories, Chemistry and Peace. While generating the dataset used in our evaluations, the annotators randomly selected items to belong to a category from the full set of available items, without taking into account occasional cases where an item may belong into two categories. This case highlights the fact that \textit{EVE} explanations are meaningful and can inform the choices made by human annotators.
\begin{table}
\caption{Sample explanation for the \textit{sorting relevant items first} task, for the query: \{Classical music\}. All top-6 features are Wikipedia categories except those beginning with `$\alpha$:' which are Wikipedia articles.
}
\label{tab:exp-sorting-music}       
\begin{tabular}{|l|l|}
\hline
\textbf{Ludwig van Beethoven (Classical)
} & \textbf{Herbie Hancock (Jazz)}\\  
\textbf{First correct found at \#1} & \textbf{First incorrect found at \#18}\\  
\hline
romantic composers & 20th-century american musicians\\ 
\hline
\textbf{19th-century classical composers} & \textbf{$\alpha$:classical music}\\ 
\hline
composers for piano & american jazz composers\\ 
\hline
\textbf{german male classical composers} & grammy award winners\\ 
\hline
german classical composers & $\alpha$:herbie hancock\\ 
\hline
19th-century german people & american jazz bandleaders\\ 
\hline
\end{tabular}
\end{table}

In Table \ref{tab:exp-sorting-music}, the query is `category' \textit{Classical music} from the ``topical type'' \textit{music genres}. We see that the first correct result is observed at rank 1 and the first incorrect result is at rank 18. The bold features show that both individuals are associated with classical music. Looking at the biography of the musician Herbie Hancock more closely, we find that he received an education in classical music and he is also well known in the classical genre, although not as strongly as he is known for Jazz music. This again goes to show that explanations generated using the \textit{EVE} model are insightful and can support the activity of manual annotators.

\section{Conclusion and Future Directions}
In this contribution, we presented a novel technique, \textit{EVE}, for generating vector representations of words using information from Wikipedia. This work represents a first step in the direction of explainable word embeddings, where the core of this interpretability lies in the use of labeled vector dimensions corresponding to either Wikipedia categories or Wikipedia articles. We have demonstrated that, not only are the resulting embeddings useful for fundamental data mining tasks, but the provision of labeled dimensions readily supports the generation of task-specific explanations via simple vector operations. 
We do not argue that embeddings generated on structured data, such as those produced by the \textit{EVE} model, would replace the prevalent existing word embedding models. Rather, we have shown that using structured data can provide additional benefits beyond those afforded by existing approaches. An interesting aspect to consider in future would be the use of hybrid models, generated on both structured data and unstructured text, which could still retain aspects of explanations as proposed in this work. 

In future, we would like to investigate the effect of the popularity of a word or concept (\ie the number of non-zero dimensions in the embedding). For example, a cuisine-related item might have fewer non-zero dimensions when compared to a country-related item. Similarly, an interesting direction might be to analyze embedding spaces and sub-spaces to learn more about correlations of dimensions, while addressing a task or the effects of dimensionality reduction (even though spaces may be sparse). Another interesting avenue for future work could be to explore different ways of generating task-specific explanations, and to investigate how these explanations might be presented effectively to a user.  

\vskip -1.6em




\vskip 0.7em
\vskip 1.5em
\noindent \textbf{Acknowledgements.} This publication has emanated from research conducted with the support of Science Foundation Ireland (SFI), under Grant Number SFI/12/ RC/2289.

%
%


\bibliographystyle{spbasic}      
\bibliography{main}   

%
%

\end{document}